\documentclass[10pt,twocolumn,letterpaper]{article}

\usepackage[pagenumbers]{cvpr} 

\usepackage{multirow}

\definecolor{cvprblue}{rgb}{0.21,0.49,0.74}
\usepackage[pagebackref,breaklinks,colorlinks,allcolors=cvprblue]{hyperref}

\def\paperID{7} 
\def\confName{CVPR}
\def\confYear{2025}

\title{PEO: Training-Free Aesthetic Quality Enhancement in Pre-Trained Text-to-Image Diffusion Models with \underline{P}rompt \underline{E}mbedding \underline{O}ptimization}

\author{
Hovhannes Margaryan$^{1}$\footnotemark[1] \quad
Bo Wan$^{1}$\footnotemark[2] \quad
Tinne Tuytelaars$^{1}$\\
$^{1}$KU Leuven
}

\newcommand{\crefT}[1]{Fig. \hyperref[#1]{\textcolor{red}{T-1}}}

\usepackage{subcaption}            

\begin{document}

\twocolumn[{%
  \renewcommand\twocolumn[1][]{#1}%
  \maketitle
  \begin{center}
    \captionsetup{type=figure, labelformat=empty}
    \begin{subfigure}[b]{0.49\textwidth}
      \includegraphics[width=\textwidth]{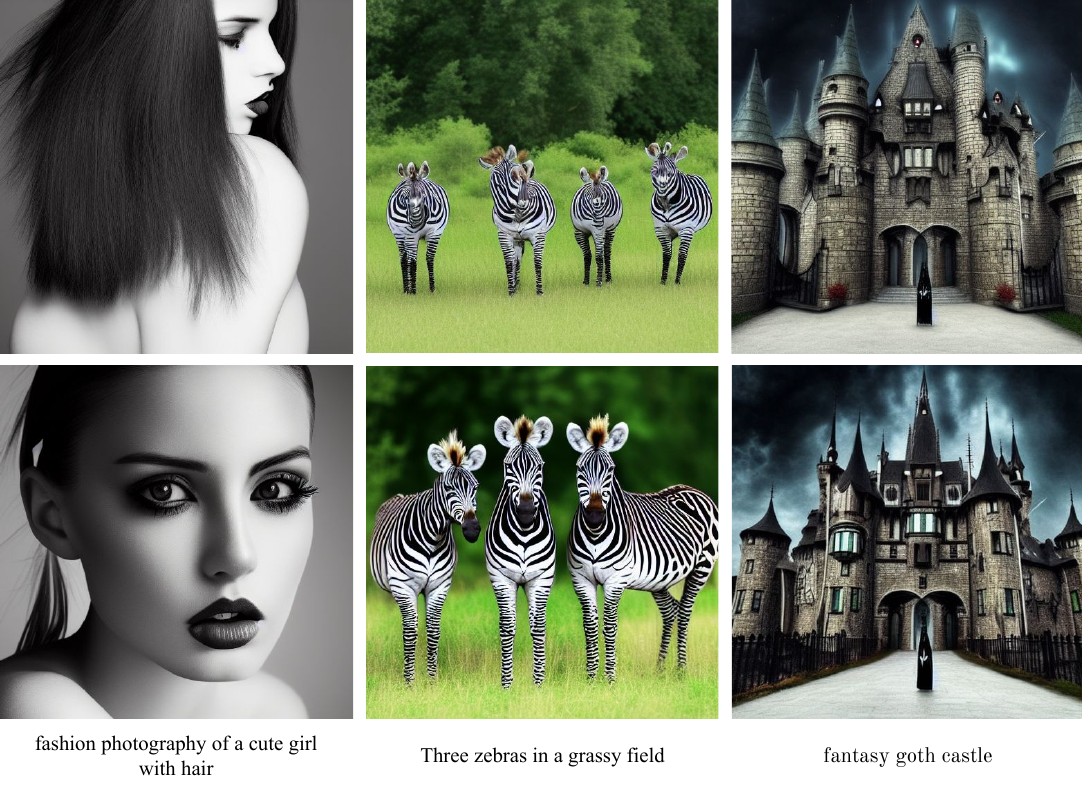}
      \caption{SD-v1-5 \cite{Rombach_2022_CVPR} without (top) and with (bottom) PEO}
    \end{subfigure}\hfill
    \begin{subfigure}[b]{0.49\textwidth}
      \includegraphics[width=\textwidth]{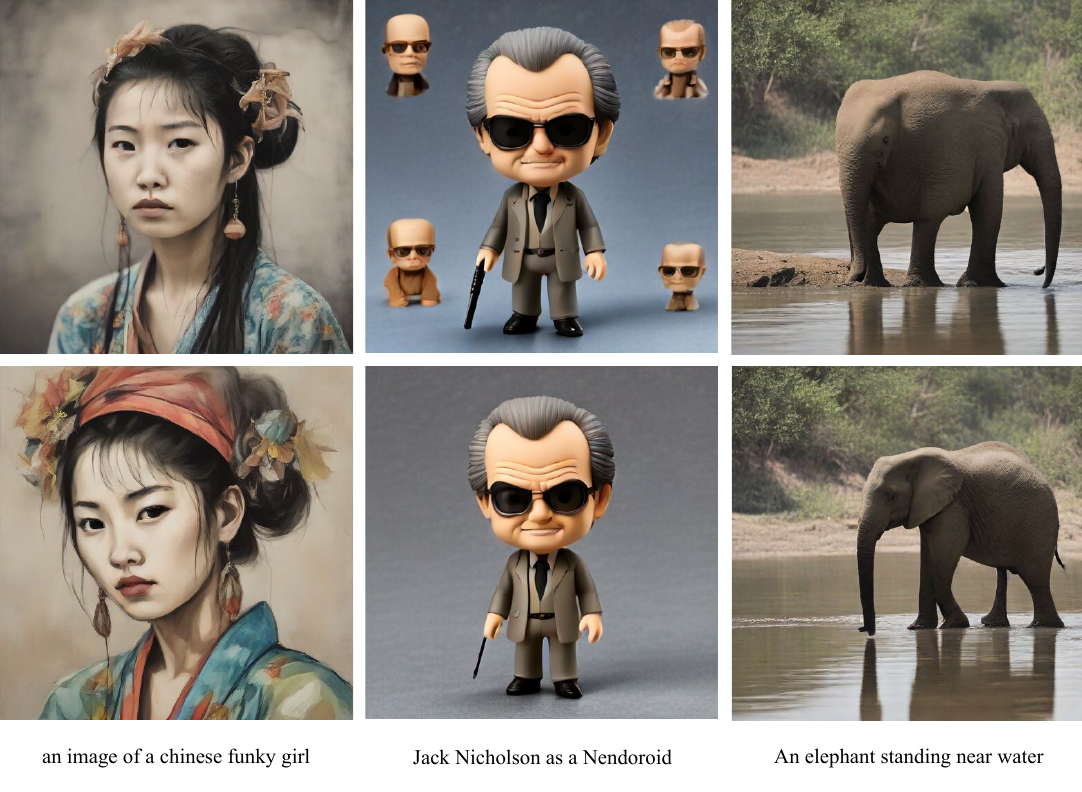}
      \caption{SDXL Turbo \cite{sauer2023adversarial} without (top) and with (bottom) PEO}
    \end{subfigure}
    \captionof{figure}{Figure T-1. Visual comparison of using initial (top) and optimized prompt embedding (bottom) in pre-trained text-to-image diffusion models. The proposed \underline{P}rompt \underline{E}mbedding \underline{O}ptimization (PEO) improves the aesthetic fidelity of the generated image both for (a) SD-v1-5 \cite{Rombach_2022_CVPR} and (b) SDXL Turbo \cite{sauer2023adversarial} and achieves adherence to the optimized text embedding and minimal deviation from the original prompt.}
    \label{fig:fig_0}
  \end{center}%
}]

\pagestyle{plain}
\thispagestyle{plain}

\begingroup
\renewcommand\thefootnote{\fnsymbol{footnote}} 
\footnotetext[1]{Work done as part of H. Margaryan's master's thesis at KU Leuven. Current affiliation: Université Paris-Saclay.}
\footnotetext[2]{Current affiliation: Meta.}
\endgroup

\begin{abstract}
    This paper introduces a novel approach to aesthetic quality improvement in pre-trained text-to-image diffusion models when given a simple prompt. Our method, dubbed Prompt Embedding Optimization (\textbf{PEO}), leverages a pre-trained text-to-image diffusion model as a backbone and optimizes the text embedding of a given simple and uncurated prompt to enhance the visual quality of the generated image. We achieve this by a tripartite objective function that improves the aesthetic fidelity of the generated image, ensures adherence to the optimized text embedding, and minimal divergence from the initial prompt. The latter is accomplished through a prompt preservation term. Additionally, PEO is training-free and backbone-independent. Quantitative and qualitative evaluations confirm the effectiveness of the proposed method, exceeding or equating the performance of state-of-the-art text-to-image and prompt adaptation methods. The code of our method is available at \href{https://github.com/marghovo/PEO}{https://github.com/marghovo/PEO}. 
\end{abstract}

\section{Introduction}
\label{sec:sec_0}

In recent years, text-to-image generative models have made notable advancements \cite{Rombach_2022_CVPR, ramesh2022hierarchical, NEURIPS2022_ec795aea, 10.5555/3692070.3692573, dai2023emuenhancingimagegeneration}, particularly due to the emergence of diffusion models \cite{pmlr-v37-sohl-dickstein15, song2021denoising, NEURIPS2020_4c5bcfec, song2021scorebased, NEURIPS2021_49ad23d1}. Text-to-image generation synthesizes an output image conditioned on a prompt. State-of-the-art text-to-image diffusion models \cite{Rombach_2022_CVPR, podell2023sdxl, sauer2023adversarial, lin2024sdxllightning} demonstrate promising results in high-fidelity image generation. However, their reliance on the complexity of the input text poses a challenge. \cref{fig:fig_1} demonstrates an example to showcase the importance of the input prompt on the quality of the generated image. On the one hand, the image generated with a simple prompt (\cref{fig:fig_1} (a)) lacks details and is not particularly appealing. On the other hand, the image generated with a carefully designed prompt (\cref{fig:fig_1} (b)) bears fine details, is pleasing to the human eye, and aligns with the given text. A simple prompt is defined as one that contains only a few words describing the image, focuses on the primary subject, and omits detailed descriptions of the subject, its attributes, and picture style. High-quality text-to-image generation from simple prompts holds applications across diverse fields including art, design, synthetic data generation, and content creation.

High-fidelity text-to-image generation using a simple prompt is challenging due to the requirements of high-level semantic understanding, fine-level details, and cross-modal alignment. Existing methods address high-quality image generation by prompt engineering at inference, using sampling guidance \cite{ho2022classifierfree, 10378223}, model retraining \cite{podell2023sdxl, li2024playground}, or prompt adaptation \cite{NEURIPS2023_d346d919}. On the one hand, crafting a prompt at inference entails a significant time to achieve a visually appealing image and can become a tedious process.  On the other hand, sampling guidance requires external information to guide the generation process (e.g. the result of an unconditional generation) and often fails when provided with an uncurated prompt. Lastly, model retraining and prompt adaptation methods require time and labor for image and prompt collection and extensive computational resources for model fine-tuning or training. To mitigate these issues, this paper proposes a novel formulation of image aesthetics improvement in pre-trained text-to-image diffusion models through text embedding optimization when given a simple prompt, called \textbf{\underline{P}rompt \underline{E}mbedding \underline{O}ptimization (PEO)}. Our method receives a simple prompt as input and optimizes its embedding to improve the aesthetic quality of the generated image by the pre-trained text-to-image diffusion model. Additionally, our approach maintains close alignment with the given simple prompt. An overview of results obtained by the proposed method is provided in \crefT{fig:fig_0}.  

PEO uses a pre-trained diffusion model (e.g. SD-v1-5 \cite{Rombach_2022_CVPR}) as a backbone and comprises an objective function that handles the following aspects of high-fidelity text-to-image generation: aesthetic quality and adherence to the prompt. First, LAION Aesthetic Predictor V2 (LAION-AesPredv2) \cite{schuhmann_improved_aesthetic_2023} is used as the first term of the objective function to obtain a visual quality score for the generated image. Second, cosine similarity is computed between the features of the generated image obtained by CLIP’s \cite{pmlr-v139-radford21a} image encoder and the text embedding being optimized to ensure adherence between them. Third, a cosine similarity between the initial text embedding and the text embedding being optimized is calculated to attain minimal deviation from the original prompt. Each term of the objective function is controlled by a hyperparameter.

\setcounter{figure}{0}
\renewcommand{\thefigure}{\arabic{figure}}

\begin{figure}
  \centering   \includegraphics[width=1.0\linewidth]{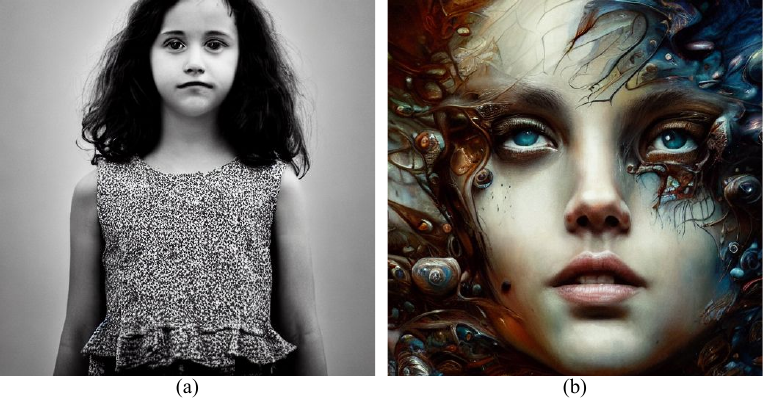}
   \caption{Two images generated with SD-v1-5 \cite{Rombach_2022_CVPR}. (a) is generated with a simple and uncurated prompt: \textit{"photo of a girl"} and lacks intricacies. (b) is generated with a carefully designed prompt: \textit{"1girl, 8k resolution, photorealistic masterpiece by Aaron Horkey and Jeremy Mann, intricately detailed fluid gouache painting by Jean Baptiste, professional photography, natural lighting, volumetric lighting, maximalist, 8k resolution, concept art, intricately detailed, complex, elegant, expansive, fantastical, cover"} and is visually appealing and highly detailed. Prompt for (b) is taken from Fotor \protect\footnotemark.}
   \label{fig:fig_1}
\end{figure}

\footnotetext{https://www.fotor.com/blog/stable-diffusion-prompts/}

The contributions of this paper are multi-fold: 

\begin{itemize}
    \item A novel simple, training-free, and backbone-independent formulation of image aesthetic quality improvement in pre-trained text-to-image diffusion models using prompt embedding optimization. 
    \item Introducing a tripartite objective function that allows visual quality improvement in the generated image, compliance with the optimized text embedding, and minimal deviation from the original prompt. The latter is accomplished by a novel \underline{P}rompt \underline{p}reservation \underline{t}erm (PPT) that guarantees an optimal text embedding stays in the neighborhood of the initial prompt embedding. 
    \item Experiments, a user study and an ablation study on prompts from DiffusionDB \cite{wangDiffusionDBLargescalePrompt2022}, COCO \cite{chen2015microsoftcococaptionsdata} and a custom set of simple captions to show the efficiency of the proposed method, surpassing or matching the state-of-the-art text-to-image and prompt adaptation methods.    
\end{itemize}

\begin{figure*}[t]
  \centering   \includegraphics[width=1.0\linewidth]{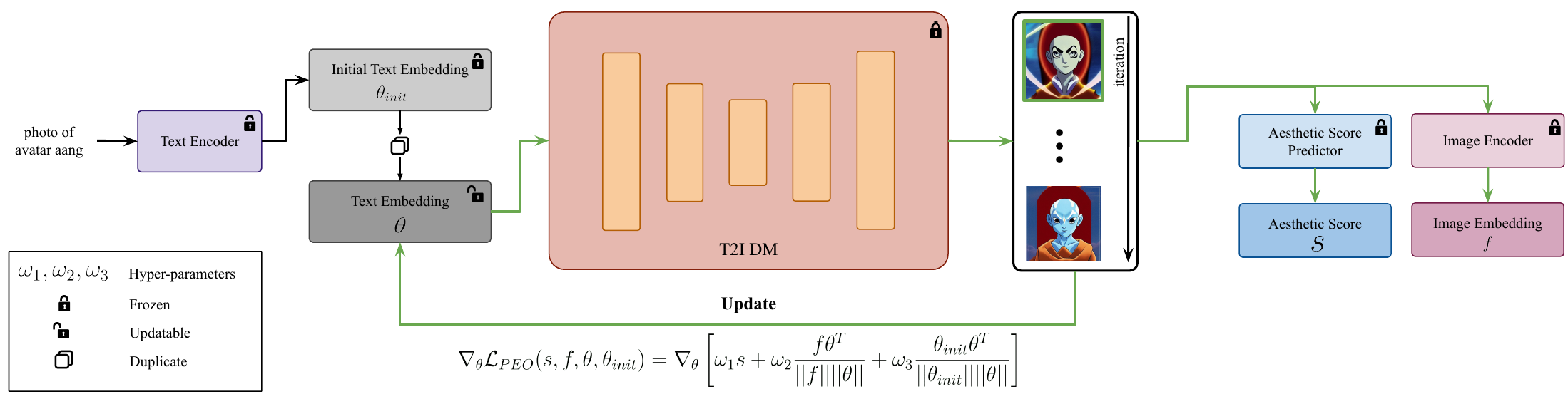}
   \caption{The optimization framework of the proposed PEO method. PEO optimizes the embedding of the given prompt using \(\mathcal{L}_{PEO}\) as an objective function which takes into account the aesthetic quality of the generated image, the distance between the text embedding being optimized and features of the generated image in CLIP's space, and the similarity between the optimized and initial text embeddings. While the visualization is in pixel space, we work in latent space.}
   \label{fig:fig_00}
\end{figure*}

\section{Related Work}
\label{sec:sec_1}

\textbf{Text-to-Image Diffusion Models (T2I DMs).} T2I DMs are a subset of conditional diffusion models that are guided by a prompt \cite{NEURIPS2022_ec795aea, ramesh2022hierarchical, Rombach_2022_CVPR, podell2023sdxl, sauer2023adversarial, lin2024sdxllightning}. Imagen \cite{NEURIPS2022_ec795aea} employs a large, pre-trained transformer language model to guide the diffusion process for generating images based on text. DALL-E 2 \cite{ramesh2022hierarchical} first generates CLIP \cite{pmlr-v139-radford21a} image embeddings based on text and then a decoder is used to generate an image given image embeddings. In contrast, the latent diffusion model \cite{Rombach_2022_CVPR} performs the diffusion process in a compressed latent space, which decreases training and inference times while maintaining high-quality image generation. SDXL \cite{podell2023sdxl} extends the denoising backbone model by additional attention modules, and two text encoders, and enables image generation at various aspect ratios by conditioning the backbone on the image size. Additionally, SDXL introduces an optional refiner model to enhance generation results through image-to-image translation. On the other hand, SDXL Turbo \cite{sauer2023adversarial} reduces the number of sampling steps required at inference using an adversarial loss for diffusion distillation in a student-teacher framework with a discriminator network. At training, both the discriminator loss, which compares real images and the output of the student network, and a distillation loss, which compares outputs from the student and teacher networks, are optimized. SDXL Turbo suffers from image blurriness in its outputs. To mitigate this, SDXL-Lightning \cite{lin2024sdxllightning} employs an adversarial objective as the distillation loss instead of an L2 loss.

\textbf{Sampling guidance.} Using only a T2I DM often fails to achieve high-fidelity results. To improve the visual quality of generated images, additional guiding techniques have been introduced \cite{NEURIPS2021_49ad23d1, ho2022classifierfree, pmlr-v162-nichol22a, 10378223}. Classifier Guidance (CG) \cite{NEURIPS2021_49ad23d1} uses class information at sampling, and controls its influence by a guidance scale. Specifically, a classifier is trained on noisy images and its gradients are used at sampling to steer the diffusion process toward a particular class. The primary issue associated with CG is the need to train an additional classifier. To circumvent this, Classifier-Free Guidance (CFG) \cite{ho2022classifierfree} uses the output of an unconditional diffusion model at inference. A single conditional diffusion model is trained by randomly dropping the conditions to facilitate unconditional generation at inference. Additionally, GLIDE \cite{pmlr-v162-nichol22a} compares two guiding techniques: CLIP guidance and CFG. In CLIP guidance, the classifier in CG is replaced by a CLIP model trained on noisy images. GLIDE demonstrates that CFG surpasses CLIP guidance in producing high-quality text-guided image generation results.

\textbf{Prompt adaptation.} Sampling guidance improves the quality of generated results compared to generation without guidance. However, it still requires carefully designed prompts to achieve high-fidelity results. To address this limitation, prompt adaptation offers automatic transformation of user-provided prompts to improve sample quality. For example, Promptist \cite{NEURIPS2023_d346d919} fine-tunes a GPT-2 \cite{noauthororeditor} model on curated prompts and then incorporates a reward function within a reinforcement learning setting to produce visually appealing images. The reward function includes LAION-AesPredv2 \cite{schuhmann_improved_aesthetic_2023} and considers the similarity between the original prompt and the generated image.

\section{Method}
\label{sec:sec_2}

This section first outlines the problem of improving the visual quality of images generated by a pre-trained T2I DM given a simple prompt using prompt embedding optimization. Second, the proposed tripartite objective function of PEO is presented along with the prompt preservation term. 

\subsection{Prompt Embedding Optimization}

The goal of PEO is to enhance the aesthetic fidelity of the generated image by a pre-trained T2I DM when provided with an uncurated prompt. Formally, given a pre-trained T2I DM \(\mathcal{G}\) (e.g. SD-v1-5 \cite{Rombach_2022_CVPR}), a simple prompt \(P\) and its corresponding \(d\)-dimensional vector representation \(\theta_{init} \in R^d\) obtained by the text encoder \(\mathcal{E_T}\) of the CLIP \cite{pmlr-v139-radford21a} model (i.e. \(\theta_{init} = \mathcal{E_T}(P)\)) PEO aims to navigate the text embedding space of the CLIP model to identify a text embedding \(\theta^*\) that improves the visual quality of the generated image relative to the image produced using \(\theta_{init}\). To this end, a tripartite objective function \(\mathcal{L}_{PEO}\) is maximized to find an optimal text embedding \(\theta^*\).

\begin{equation}
    \label{eq:eq_10}
    \theta^{*} = {\arg \max}_{\theta} [\mathcal{L}_{PEO}].
\end{equation}

\begin{figure*}[t]
  \centering   
  \includegraphics[width=1.0\linewidth]{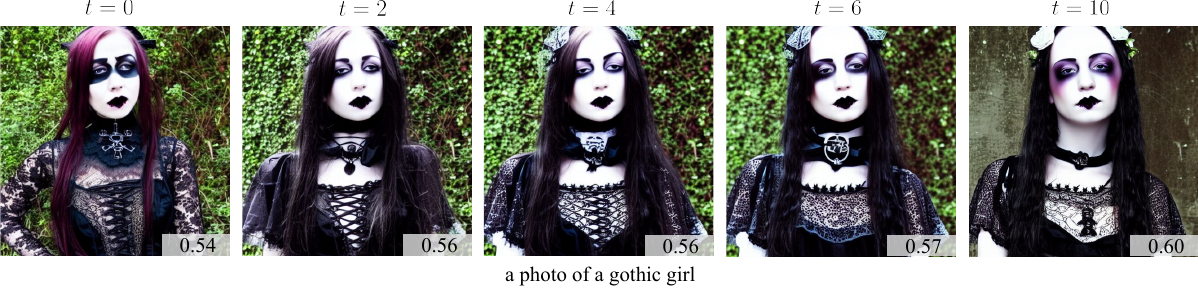}
   \caption{The proposed PEO optimization method in image space, with \(t\) as the current step. At \(t=10\), the aesthetic quality of the generated image and text-to-image relevance are improved compared to \(t=0\). The bottom-right number is the LAION-AesPredv2 value.}
   \label{fig:fig_9}
\end{figure*}

The pipeline of the proposed method is presented in \cref{fig:fig_00}. Given a pre-trained T2I DM and an initial simple prompt, the former's embedding is first used to generate an image conditioned on it. Secondly, the value of the objective function is computed for the generated image and the corresponding text embedding. Finally, the text embedding is updated using the gradients of the objective function. \cref{fig:fig_9} illustrates the optimization procedure of PEO over ten steps in image space. After each optimization step, the current text embedding is used to generate an image.

\subsection{Tripartite Objective Function}

The proposed objective function, which includes three terms, aims to enhance the aesthetic quality of the generated image, maintain fidelity to the optimized text embedding, and minimize the divergence of \(\theta^*\) from the initial prompt, thereby addressing key aspects of high-quality text-to-image generation. Inspired by Promptist \cite{NEURIPS2023_d346d919}, pre-trained LAION-AesPredv2 \cite{schuhmann_improved_aesthetic_2023}, \(\mathcal{S}\), is used to compute a visual quality score for the generated image: \(s = \mathcal{S(\mathcal{G}(\theta))}\), where \(\theta\) represents the text embedding under optimization. Thus, the first component of the objective is \(\mathcal{L}_1 = \mathcal{S(\mathcal{G}(\theta))} = s\).
By default, \(\mathcal{S}\) predicts a human preference score ranging from \(0\) to \(10\); in practice, we normalize this score to a range of \(0\) to \(1\).
 
The second term of \(\mathcal{L}_{PEO}\) is formulated as:

\begin{equation}
    \label{eq:eq_7}
     \mathcal{L}_2 = \frac{f\theta^T}{||f|| ||\theta||},
\end{equation}
 
where \(f=\mathcal{E_I}(\mathcal{G(\theta)})\) are the image features of the generated image, obtained by the image encoder \(\mathcal{E_I}\) of CLIP. \cref{eq:eq_7} computes the cosine similarity between the features of the generated image and the text embedding being optimized. \(\mathcal{L}_2\) aims to ensure adherence between the generated image and the optimized text embedding.

\textbf{Prompt Preservation Term:} The \underline{p}rompt \underline{p}reservation \underline{t}erm (PPT) of the proposed objective function is:

\begin{equation}
    \label{eq:eq_8}
     \mathcal{L}_{PPT} = \frac{\theta_{init} \theta^T}{||\theta_{init} || ||\theta||}.
\end{equation}

PPT computes a cosine similarity between the initial text embedding \(\theta_{init}\) and the text embedding being optimized \(\theta\). Hence, it ensures minimal deviation of the optimized text embedding from the initial text embedding. Experiments show that omitting this term from the objective function leads to divergence from the original prompt and a loss of the identity specified in the prompt. \cref{sec:sec_4_6} provides an ablation study on the objective function's terms.

The tripartite objective function is then defined as:

\begin{equation}
    \label{eq:eq_9}
    \mathcal{L}_{PEO}(s, f, \theta, \theta_{init}) = \omega_1 \mathcal{L}_1 + \omega_2
    \mathcal{L}_2
    + \omega_3 \mathcal{L}_{PPT}
\end{equation}

where \(\omega_1\), \(\omega_2\), \(\omega_3\) are hyperparameters controlling the influence of each term. A hyperparameter search is conducted in the appendix to demonstrate the influence of each coefficient on the generated output.

Upon completion of the prompt embedding optimization (i.e., when the maximum number of optimization steps is reached or the objective function value no longer increases), an optimal text embedding \(\theta^{*}\) is obtained. This embedding is then used to generate an image that is aesthetically more pleasing than the one produced with \(\theta_{init}\). Additionally, the generated image adheres to \(\theta^{*}\), and   \(\theta^{*}\) remains close to \(\theta_{init}\) due to PPT, ensuring that the generated image accurately represents the provided simple prompt. In practice, we integrate PEO with classifier-free guidance. Notably, only the text embedding of the provided prompt is optimized, while the text embedding used for unconditional generation remains unchanged.

\begin{table*}
\centering
\begin{tabular}{c c c c | c c c}
\hline
\multicolumn{7}{c}{\textbf{Using SD-v1-5 as a backbone}} \\ \hline
\multirow{4}{*}{} & \multicolumn{3}{c|}{DiffusionDB} & \multicolumn{3}{c}{COCO} \\ 
                  & SD-v1-5 & Promptist & Ours & SD-v1-5 & Promptist & Ours \\ \hline
LAION-AesPredv2 ↑ & 0.58 ± 0.0033 & \textbf{0.64} ± 0.0045 & \underline{0.61} ± 0.0026 & 0.64 ± 0.0016 & \textbf{0.64} ± 0.0041 & \underline{0.59} ± 0.0013 \\ 
HPSv2 ↑           & 0.26 ± 0.0002   & 0.26 ± 0.0002   & 0.26 ± 0.0002   & 0.27 ± 0.0002   & 0.27 ± 0.0002   & 0.27 ± 0.0002 \\ 
CLIPScore ↑       & \textbf{0.28} ± 0.0020 & 0.27 ± 0.0025 & \textbf{0.28} ± 0.0022 & \textbf{0.26} ± 0.0011 & 0.25 ± 0.0013 & \textbf{0.26} ± 0.0009 \\ \hline
& \multicolumn{3}{c|}{DiffusionDB (simplified)} & \multicolumn{3}{c}{COCO (simplified)} \\ 
& SD-v1-5 & Promptist & Ours & SD-v1-5 & Promptist & Ours \\ \hline
LAION-AesPredv2 ↑ & 0.57 ± 0.0028 & \textbf{0.63} ± 0.0040 & \underline{0.60} ± 0.0020 & 0.57 ± 0.0016 & \textbf{0.64} ± 0.0034 & \underline{0.59} ± 0.0014 \\ 
HPSv2 ↑           & 0.25 ± 0.0002   & 0.25 ± 0.0002   & \textbf{0.26} ± 0.0002      & \textbf{0.27} ± 0.0002 & 0.26 ± 0.0001   & \textbf{0.27} ± 0.0002 \\ 
CLIPScore ↑       & \textbf{0.27} ± 0.0016 & 0.26 ± 0.0022 & \textbf{0.27} ± 0.0017 & \textbf{0.26} ± 0.0011 & 0.25 ± 0.0001 & \textbf{0.26} ± 0.0010 \\ \hline \\[2pt]\hline

\multicolumn{7}{c}{\textbf{Using SDXL Turbo as a backbone}} \\ \hline
\multirow{4}{*}{} & \multicolumn{3}{c|}{DiffusionDB} & \multicolumn{3}{c}{COCO} \\ 
                  & SDXL Turbo & Promptist & Ours & SDXL Turbo & Promptist & Ours \\ \hline 
LAION-AesPredv2 ↑ & 0.66 ± 0.0039 & \textbf{0.70} ± 0.0036 & \underline{0.68} ± 0.0036 & 0.57 ± 0.0018 & \textbf{0.69} ± 0.0054 & \underline{0.59} ± 0.0018 \\ 
HPSv2 ↑           & 0.27 ± 0.0002   & 0.27 ± 0.0002   & 0.27 ± 0.0002   & \textbf{0.28} ± 0.0002   & 0.27 ± 0.0002   & \textbf{0.28} ± 0.0002 \\ 
CLIPScore ↑       & \textbf{0.29} ± 0.0018 & 0.28 ± 0.0024 & \textbf{0.29} ± 0.0020 & \textbf{0.27} ± 0.0010 & 0.26 ± 0.0013 & \textbf{0.27} ± 0.0012 \\ \hline
& \multicolumn{3}{c|}{DiffusionDB (simplified)} & \multicolumn{3}{c}{COCO (simplified)} \\ 
& SDXL Turbo & Promptist & Ours & SDXL Turbo & Promptist & Ours \\ \hline
LAION-AesPredv2 ↑ & 0.66 ± 0.0038 & \textbf{0.71} ± 0.0033 & \underline{0.68} ± 0.0035 & 0.58 ± 0.0021 & \textbf{0.71} ± 0.0041 & \underline{0.60} ± 0.0020 \\ 
HPSv2 ↑           & 0.27 ± 0.0002   & 0.27 ± 0.0002   & 0.27 ± 0.0002   & \textbf{0.28} ± 0.0002   & 0.27 ± 0.0002   & \textbf{0.28} ± 0.0002 \\ 
CLIPScore ↑       & \textbf{0.28} ± 0.0016 & 0.27 ± 0.0019 & \textbf{0.28} ± 0.0016 & 0.27 ± 0.0010 & 0.27 ± 0.0009 & 0.27 ± 0.0011 \\ \hline
\end{tabular}
\caption{Quantitative comparison among the baselines and the proposed method on DiffusionDB, COCO, DiffusionDB (simplified), and COCO (simplified) datasets using SD-v1-5 and SDXL Turbo as backbones. Our method surpasses or matches the baselines in aesthetic quality without compromising text-to-image relevance.}
\label{tab:tab_1}
\end{table*}


\begin{table*}
\centering
\begin{tabular}{c c c c | c c c}
\hline
\multicolumn{7}{c}{PEO Dataset} \\ \hline
 & \multicolumn{3}{c|}{Using SD-v1-5 as a backbone} & \multicolumn{3}{c}{Using SDXL Turbo as a backbone} \\ 
 & SD-v1-5 & Promptist & Ours & SDXL Turbo & Promptist & Ours \\ \hline
LAION-AesPredv2 ↑ & 0.60 ± 0.0022 & \textbf{0.64} ± 0.0020 & \underline{0.63} ± 0.0030 & 0.66 ± 0.0041 & \textbf{0.70} ± 0.0036 & \underline{0.68} ± 0.0043 \\
HPSv2 ↑           & 0.26 ± 0.0003 & 0.26 ± 0.0003 & 0.26 ± 0.0003 & 0.27 ± 0.0003 & 0.27 ± 0.0003 & 0.27 ± 0.0003 \\
CLIPScore ↑       & 0.26 ± 0.0011 & 0.25 ± 0.0011 & 0.25 ± 0.0018 & 0.28 ± 0.0013 & 0.27 ± 0.0011 & \textbf{0.28} ± 0.0014 \\ \hline
\end{tabular}
\caption{Quantitative comparison among the baselines and the proposed method on the PEO dataset using SD-v1-5 and SDXL Turbo as backbones. Our method outperforms or matches the baselines in aesthetic quality while maintaining text-to-image alignment.}
\label{tab:tab_2}
\end{table*}

\section{Experiments}
\label{sec:sec_3}

\subsection{Dataset}
\label{sec:sec_3_1}

Experiments of the proposed method are conducted on the same subset of DiffusionDB \cite{wangDiffusionDBLargescalePrompt2022} (256 prompts)  and COCO \cite{chen2015microsoftcococaptionsdata} (200 prompts) datasets as used for evaluation in Promptist \cite{NEURIPS2023_d346d919}. To align with the aim of PEO (i.e. improving aesthetic quality of the generated image from simple prompts), we apply GPT-4 \cite{openai2023gpt4} to make the prompts from these datasets simple using "Given the following list of prompts, make them short, focus on the main subject of the prompt." as a query. We refer to these prompt sets as "simplified." Additionally, a dataset of simple prompts is created by the authors and generated by GPT-4 with the input: "List simple prompts to test my T2I method." This dataset, referred to as the PEO dataset, contains 100 prompts and covers a wide range of objects, scenes, and styles. All sets of prompts are included in the appendix.

\subsection{Implementation and Evaluation Metrics}

\begin{figure*}[htbp!]
  \centering   \includegraphics[width=1.0\linewidth]{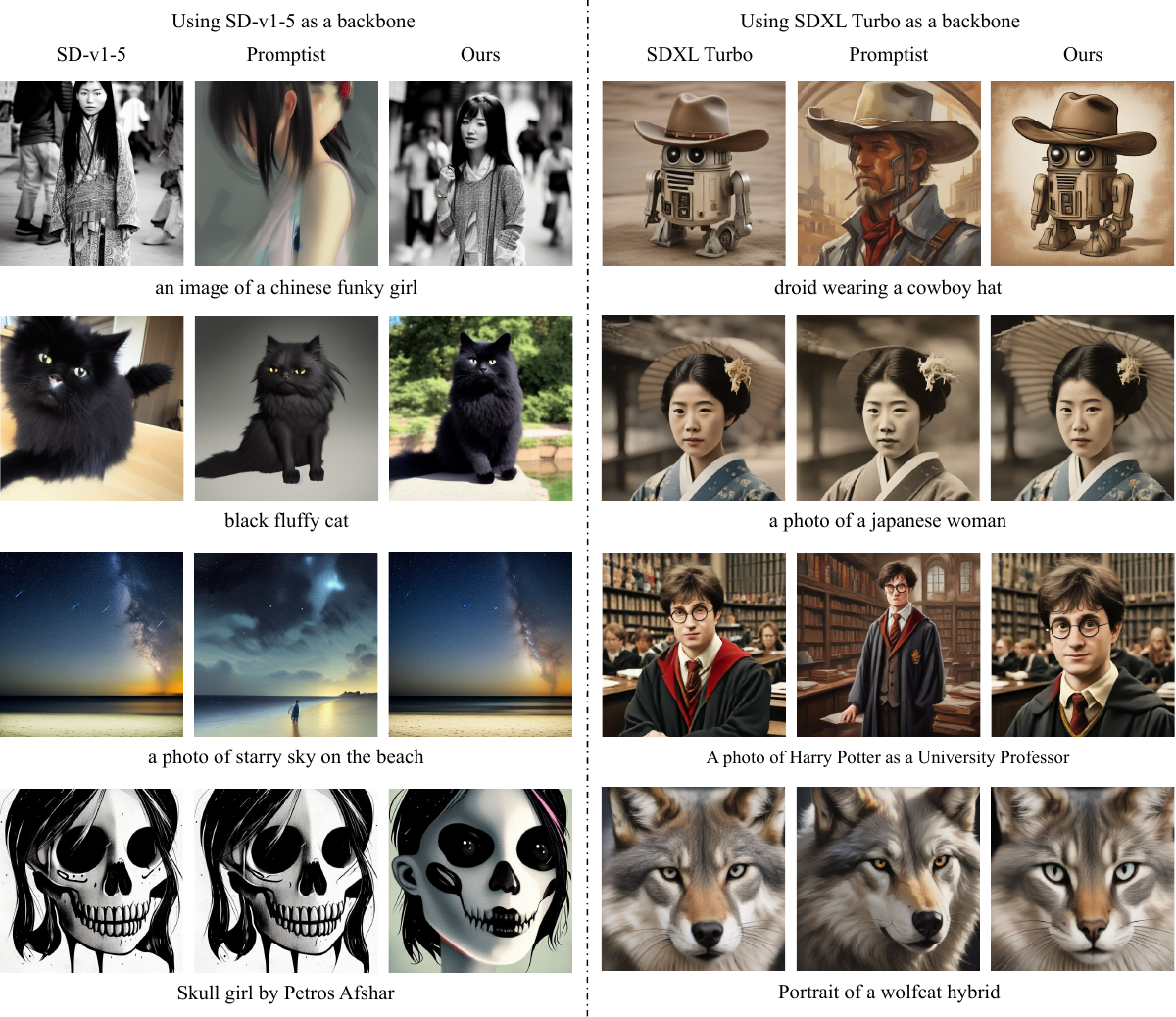}
   \caption{Qualitative comparison of PEO and baselines with SD-v1-5 and SDXL Turbo as backbones. Our approach surpasses the baseline in visual aesthetic quality, exhibits improved details, and better alignment with the style and main subject indicated in the original prompt.}
   \label{fig:fig_4}
\end{figure*}

In all our experiments, PEO uses the given initial prompt with a maximum of 10 iterations and the Adam optimizer \cite{KingBa15} with \(0.01\) learning rate. As SDXL Turbo employs two text encoders we optimize text embeddings obtained from both of these encoders for the given prompt when using SDXL Turbo as a backbone. Additionally, the coefficients of the objective function terms are \(\omega_1=1.0\), \(\omega_2=0.5\), \(\omega_3=0.5\). For the backbone model (SD-v1-5 or SDXL Turbo), the initial text prompt is used to generate an image. For Promptist, the initial text prompt is provided to their model, and the output prompt is used to generate an image using the backbone. We use the UniPC scheduler \cite{zhao2023unipc} with \(15\) sampling steps and a guidance scale of \(7.5\) when SD-v1-5 serves as the backbone, and with \(1\) sampling step and a guidance scale of \(0.0\) when SDXL Turbo is the backbone.

The metrics used for automatic evaluation are LAION-AesPredv2 \cite{schuhmann_improved_aesthetic_2023}, HPSv2 \cite{wu2023human}, and CLIPScore \cite{hessel2022clipscore}. LAION-AesPredv2 evaluates visual quality, considering human preference and it is normalized to a range of \(0\) to \(1\). The inclusion of HPSv2 is strategic as relying solely on LAION-AesPredv2 for aesthetic quality assessment of the generated images may not fully demonstrate fairness, as it is inherently optimized. HPSv2 takes into account both image aesthetic quality and text-to-image relevance. CLIPScore assesses text-to-image alignment. HPSv2 and CLIPScore are computed between the initial prompt and the generated image. All random seeds are fixed for a fair comparison.

\begin{figure}
  \centering   \includegraphics[width=1.0\linewidth]{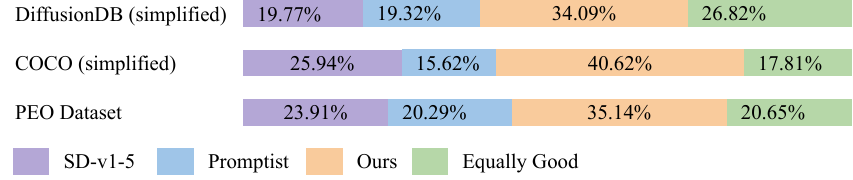}
   \caption{Results of the user study. The annotators favored PEO by at least 11.23\% over SD-v1-5 and 9.85\% over Promptist.}

   \label{fig:fig_5}
\end{figure}

\begin{figure}[htbp!]
  \centering   \includegraphics[width=1.0\linewidth]{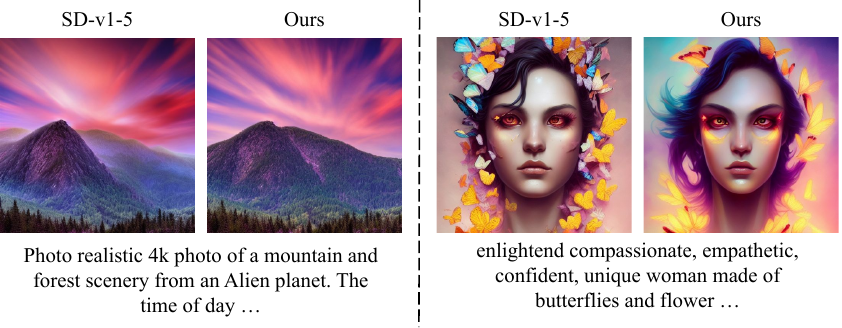}
   \caption{Visual comparison between SD-v1-5 and the proposed PEO method using hand-engineered prompts. Our method handles complex prompts and achieves a slight enhancement in visual quality over the baseline. The prompts are sourced from Fotor \protect\footnotemark.}
   \label{fig:fig_10}
\end{figure}

\footnotetext{https://www.fotor.com/blog/stable-diffusion-prompts/}

\begin{figure}[htbp!]
  \centering   \includegraphics[width=1.0\linewidth]{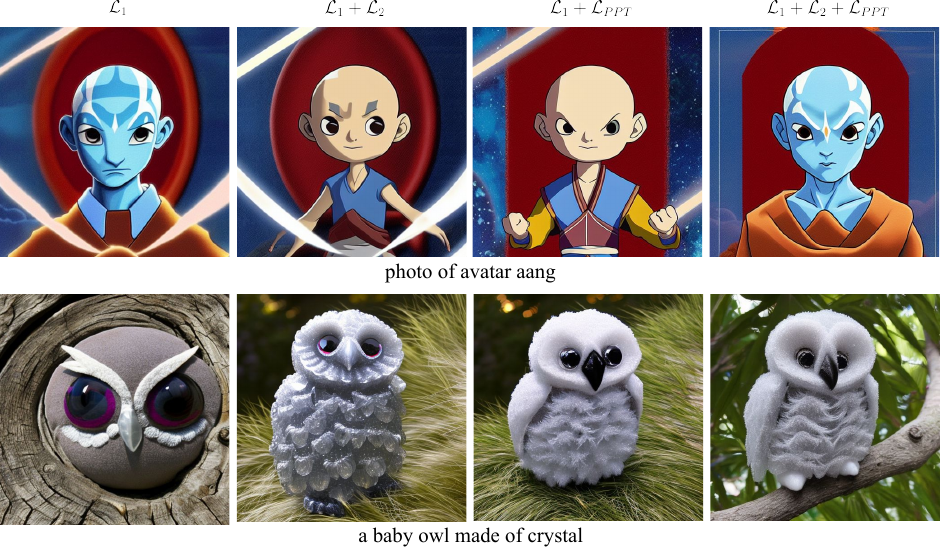}
   \caption{Qualitative comparison of different combinations of the objective function terms. The combination of all three terms results in enhancement in aesthetic quality and better alignment with the given prompt compared to other combinations.}
   \label{fig:fig_7}
\end{figure}

\subsection{Comparison with Baselines}
\label{sec:sec_3_2_1}

This section presents a quantitative and qualitative analysis and a user study comparing the proposed PEO approach with text-to-image diffusion models SD-v1-5 \cite{Rombach_2022_CVPR} and SDXL Turbo \cite{sauer2023adversarial} and the prompt adaptation technique, Promptist \cite{NEURIPS2023_d346d919} using SD-v1-5 and SDXL Turbo as backbones. Experiments with SDXL Turbo are conducted to demonstrate that PEO is backbone-independent.  

\cref{tab:tab_1,tab:tab_2} present a quantitative comparison between our method and the baselines on captions from DiffusionDB, COCO, simplified DiffusionDB, simplified COCO, and the PEO dataset using SD-v1-5 and SDXL Turbo as backbones. The scores are computed for each generated image and average scores are reported along with the variance. The proposed PEO method surpasses the backbone SD-v1-5 in LAION-AesPredv2 and achieves a similar HPSv2 score in comparison to baseline methods and outperforms Promptist in CLIPScore. Thus, our method improves the aesthetic quality of the generated image without comprising text-to-image alignment.

\cref{fig:fig_4} shows a visual comparison between the proposed method and the baselines using SD-v1-5 and SDXL Turbo as backbones on a subset of captions from the five datasets mentioned in \cref{sec:sec_3_1}. Additional results are included in the appendix. The initial text prompt is displayed below each result. The optimized prompts are not shown, as the authors are unaware of any model that can map optimized text embeddings back to text. PEO exhibits superior visual and aesthetic quality (e.g., "an image of a chinese funky girl") compared to the baselines, with enhanced detail. Additionally, PEO more accurately reflects the style indicated in the initial prompt (e.g., "funky" in "an image of a chinese funky girl"). Compared to Promptist, PEO more effectively captures the main subject of the image (e.g., "beach" in "starry sky on the beach"). While results generated by Promptist sometimes adopt a fantasy-oriented approach, they often deviate from the original prompt and the identity specified, whereas the proposed method consistently maintains realism and alignment with the given prompt. For example, in "A photo of Harry Potter as a University Professor," our method not only improves the aesthetic quality compared to SDXL Turbo but also retains the identity of "Harry Potter" better than Promptist. We hypothesize that this phenomenon is due to the PPT term included in the objective function of PEO, which helps maintain minimal divergence from the initial user prompt. It can be noticed that SDXL Turbo generates results with decent visual quality and our method still showcases improvements over the baselines. When SDXL Turbo’s results lack
detail, PEO enhances aesthetic quality and text-to-image relevance. For example, PEO improves details in the “Portrait of a wolfcat hybrid,” making the creature in the image more closely resemble a "wolfcat" compared to the baseline methods.

\textbf{User Study.} The results of the human preference evaluation are provided in \cref{fig:fig_5}. The study is conducted on 100 prompts: 44 from simplified DiffusionDB, 33 from simplified COCO, and 23 from the PEO dataset. Images are generated using the baseline methods SD-v1-5, Promptist, and PEO. Participants are asked to rank a set of generated images based on the following criteria: (1) overall aesthetic quality and realism of the image and (2) text-to-image alignment: how well the image represents the prompt in terms of style, identity, and other relevant factors. The image order is randomized for each prompt. Participants are offered the choice to select one of the methods as superior or to indicate that the results are all equally good. Ten annotators participated in our study. The scores in \cref{fig:fig_5} are shown as percentages, averaged across prompts per dataset and participant. Our method is preferred by at least 11.23\% more than SD-v1-5 and 9.85\% more than Promptist. 

Thus, our method performs similarly to the baseline based on automatic metrics. Meanwhile, the user study shows a strong preference for PEO. We hypothesize that this phenomenon arises from the limitations of existing evaluation metrics, which may fail to comprehensively capture the inherent complexities and multifaceted nature of text-to-image generation.

\textbf{Complex Prompts.} \cref{fig:fig_10} presents a visual comparison between SD-v1-5 and PEO given curated prompts to verify that the proposed method is also effective with hand-engineered prompts. When using curated prompts, the visual results generated by SD-v1-5 exhibit high fidelity, and our method manages to achieve a slight improvement over the baseline without negative visual impact.

\begin{table*}[htbp!]
  \centering
  \begin{tabular}{c c c c c c}
    \hline & \(\mathcal{L}_1\) & \(\mathcal{L}_1 + \mathcal{L}_2\) & \(\mathcal{L}_1 +\mathcal{L}_{PPT}\) & \(\mathcal{L}_1+\mathcal{L}_2+\mathcal{L}_{PPT}\) \\ \hline
    LAION-AesPredv2 ↑ & 0.60 ± 0.0021 & 0.59 ± 0.0020 & 0.60 ± 0.0023 & \textbf{0.60} ± 0.0021 \\
    HPSv2 ↑  & 0.26 ± 0.0002 & 0.26 ± 0.0002 & 0.26 ± 0.0002 & \textbf{0.27} ± 0.0002 \\
    CLIPScore ↑ & 0.27 ± 0.0017 & 0.27 ± 0.0017 & 0.27 ± 0.0017 & \textbf{0.27} ± 0.0017 \\
    \hline
  \end{tabular}
  \caption{Quantitative comparison among different combinations of the objective function terms on 150 random prompts from Promptist's \cite{NEURIPS2023_d346d919} training set. The combination of all three terms surpasses other combinations.}
  \label{tab:tab_3}
\end{table*}

\begin{figure*}[htbp!]
  \centering   \includegraphics[width=1.0\linewidth]{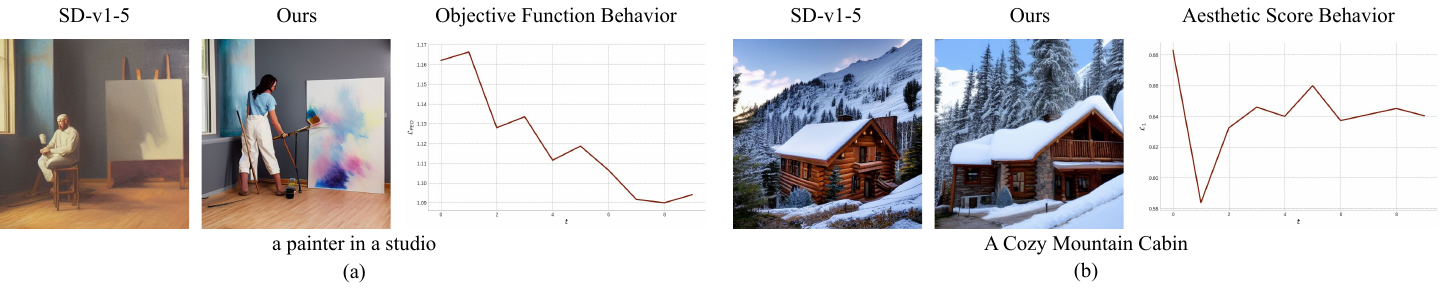}
   \caption{Failure cases of PEO. (a) Due to divergence of the optimization process. This is observed both visually in the generated results and in the behavior of the objective function. (b) Due to the backbone model generating an image with a high aesthetic score (above \(0.65\) in the case of SD-v1-5). In such cases, the proposed method may not be able to further improve the aesthetic score, as the backbone inherently cannot generate an image with an aesthetic score above a certain cutoff.}
   \label{fig:fig_14}
\end{figure*}

\subsection{Ablation Study}
\label{sec:sec_4_6}

This section presents an ablation study on the impact of each term in the objective function (\cref{eq:eq_9}) of PEO, using a random sample of 150 prompts (maximum 8 words each) from the training set released by Promptist \cite{NEURIPS2023_d346d919}. \cref{tab:tab_3} provides a quantitative comparison between different combinations of objective function terms. The combination of all three terms outperforms other combinations in quantitative metrics. Additionally, \cref{fig:fig_7} displays qualitative results of images generated with different combinations of objective function terms. Using only the first term results in a loss of identity for "avatar aang" in "photo of avatar aang", while incorporating the second and third terms ensures that the generated output adheres to the prompt and simultaneously achieves better visual quality than the other combinations.

\subsection{Failure Cases}

This section discusses the failure cases of the proposed method. First, when the optimization of the text embedding diverges. \cref{fig:fig_14} (a) demonstrates such a case with a comparison to SD-v1-5 and shows how the objective function’s value increases over the optimization steps, moving away from a local optimum. This is also visually noticeable in the generated image by PEO, where no improvement in aesthetic quality or text-to-image relevance is observed. We hypothesize that the divergence in the optimization problem occurs due to the non-convex optimization landscape of prompt embedding optimization, which has multiple local optima, making it difficult for PEO to converge.

A second scenario where PEO fails is when the generated image by the backbone model (e.g., SD-v1-5) already has a high aesthetic score (above \(0.65\), with the aesthetic score normalized to the range \(0\) to \(1\)). In such cases, our method may struggle to further maximize this score. \cref{fig:fig_14} (b) shows a visual result for such a case, comparing it with SD-v1-5 and the behavior of the aesthetic score over the optimization steps. The image generated by the backbone is of high visual quality, and the results generated by the proposed method are aesthetically similar to them. This occurs because SD-v1-5 is fine-tuned on the Laion dataset \cite{NEURIPS2022_a1859deb}, after it has been filtered by LAION-AesPredv2 with a threshold of \(0.5\). Only \(2\%\) of this dataset contains images with an aesthetic score above \(0.6\). Thus, inherently, SD-v1-5 cannot generate an image with an aesthetic score above a certain threshold. While it is not trivial to determine this threshold, our experiments reveal that if the initial text embedding produces images with a high aesthetic score as in \cref{fig:fig_14} (b), our method might be unable to maximize this score further.
\section{Conclusion}

This work presented a training-free and backbone-independent prompt embedding optimization method, PEO, that enhances the aesthetic quality of images generated from simple prompts in pre-trained text-to-image diffusion models. Given an uncurated prompt, PEO optimizes its text embedding through a tripartite objective function. The latter improves the fidelity of the generated image, ensures compliance with the optimized prompt embedding, and minimizes divergence from the original text using a novel prompt preservation term. PEO showed superior or comparable results to state-of-the-art text-to-image and prompt adaptation methods validated qualitatively and quantitatively.

\textbf{Acknowledgement.} We thank the authors of Promptist \cite{NEURIPS2023_d346d919} for providing the dataset subsets from DiffusionDB \cite{wangDiffusionDBLargescalePrompt2022} and COCO \cite{chen2015microsoftcococaptionsdata} used in their evaluation.

{\small
\bibliographystyle{ieeenat_fullname}
\bibliography{main}
}

\clearpage
\appendix
\section*{Appendix}
\label{sec:supplementary}

This supplementary material provides details on the implementation and additional results of the proposed \underline{P}rompt \underline{E}mbedding \underline{O}ptimization (PEO) method, which enhances the aesthetic quality of images generated by pre-trained text-to-image diffusion models from simple prompts. \cref{sec:sup_1} discusses the selection of the optimization algorithm and includes results from hyperparameter searches of the coefficients of the PEO objective function terms and the learning rate of the optimization algorithm. Additional visual comparison results with baseline text-to-image models (SD-v1-5 \cite{Rombach_2022_CVPR} and SDXL Turbo \cite{sauer2023adversarial}) and the prompt adaptation method, Promptist \cite{NEURIPS2023_d346d919}, are provided in \cref{sec:sup_3} using SD-v1-5 and SDXL Turbo as backbones.

\section{Choice of Optimization Algorithm and Hyperparameter Searches}
\label{sec:sup_1}

This section first explores the choice of the optimization algorithm. Second, the results of the hyperparameter search for the coefficients of the objective function terms of the proposed PEO method are provided. Finally, hyperparameter search for the learning rate of the optimization method is presented.

\subsection{Choice of Optimization Algorithm}

\begin{figure}[htbp!]
  \centering   \includegraphics[width=1.0\linewidth]{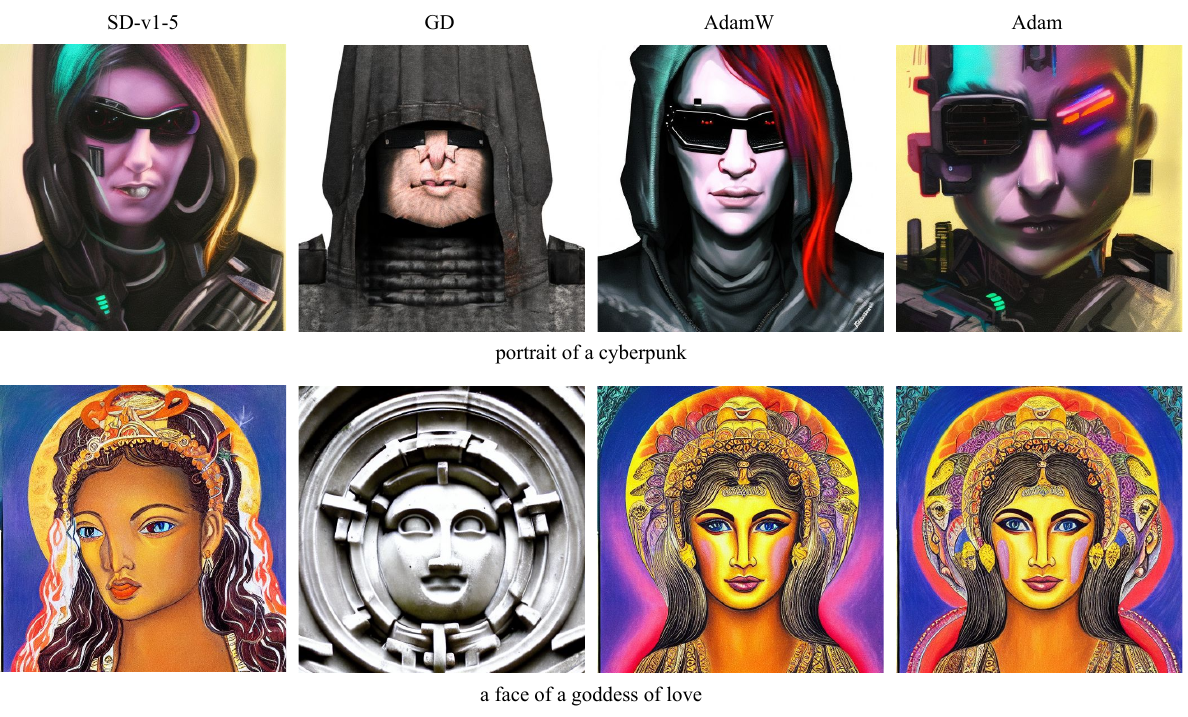}
   \caption{Visual comparison among different optimization algorithms used in the proposed method and SD-v1-5. Using Adam, the proposed method generates images that are more aesthetically pleasing and demonstrate better text-to-image relevance in comparison to other optimization methods. In this experiment, the coefficients of the objective terms are \(\omega_1=1.0\), \(\omega_2=1.0\), \(\omega_3=1.0\).}
   \label{fig:fig_3_sup}
\end{figure}

The following optimization algorithms are experimented with: Gradient Descent (GD), AdamW \cite{DBLP:conf/iclr/LoshchilovH19}, and Adam \cite{KingBa15}. \Cref{fig:fig_3_sup} and \Cref{tab:tab_7_sup} provide visual and quantitative comparisons of the proposed method using these optimization algorithms. In qualitative comparison (conducted on a random sample of 150 prompts (with a maximum of 8 words per prompt) from the training set of Promptist), it is observed that when using Adam, our method produces more aesthetically pleasing images with better relevance to the prompt than other optimization methods. \Cref{fig:fig_3_sup} shows that GD diverges and produces corrupted results because it does not utilize historical information about the gradients, unlike Adam and AdamW. Adam outperforms the other two optimization algorithms in terms of HPSv2. All three optimization methods result in a similar CLIPScore. Adam and AdamW achieve an analogous LAION-AesPredv2 which is higher than what GD achieves. Thus, we recommend using Adam as the default optimizer with the proposed method.

\begin{table*}[htbp!]
  \centering
  \begin{tabular}{c c c c c}
    \hline 
    & SD-v1-5 & GD & AdamW & Adam \\ \hline
    LAION-AesPredv2 ↑ & 0.57 ± 0.0024 & 0.59 ± 0.0022 & 0.60 ± 0.0021 & \textbf{0.60} ± 0.0022 \\
    HPSv2 ↑    & 0.26 ± 0.0002 & 0.26 ± 0.0002 & 0.26 ± 0.0002 & \textbf{0.27} ± 0.0003 \\
    CLIPScore ↑  & 0.27 ± 0.0018 & 0.27 ± 0.0020 & 0.27 ± 0.0018 & \textbf{0.27} ± 0.0019 \\
    \hline
  \end{tabular}
  \caption{Quantitative comparison among different optimization algorithms used in the proposed method and the baseline. Adam achieves higher HPSv2 and CLIPScore values compared to other optimization methods and the baseline. In this experiment, the coefficients of the objective terms are \(\omega_1=1.0\), \(\omega_2=1.0\), \(\omega_3=1.0\).}
  \label{tab:tab_7_sup}
\end{table*}

\subsection{Hyper-parameter Search for Objective Function’s Terms}

The choice of coefficients \(\omega_1\), \(\omega_2\) and \(\omega_3\) is investigated using a greedy search. Each coefficient value is sampled from the discrete set \([0.2, 0.5, 0.7, 1]\). \Cref{tab:tab_1_sup} presents a quantitative comparison among different combinations of coefficient values on a random sample of 150 prompts (with a maximum of 8 words per prompt) from the training set of Promptist. The combination \(\omega_1=1.0\), \(\omega_2=0.5\), \(\omega_3=0.5\) outperforms other combinations in HPSv2. We hypothesize that this is because the proposed \underline{P}rompt \underline{P}reservation \underline{T}erm (PPT) acts as a regularization mechanism in the objective function by minimizing the divergence of the optimal text embedding from the initial text embedding. Therefore, higher values for \(\omega_3\) compel the proposed method to produce text embeddings that remain close to the initial embedding, whereas lower values of \(\omega_3\) allow more flexibility for the method to explore text embeddings that might deviate from the original prompt. The setting of \(\omega_3\) balances the relevance of the generated image to the given prompt against the method's freedom to vary the embedding. Consequently, we observe a higher value for HPSv2, which assesses human preference based on both the image and the prompt, for the combination \(\omega_1=1.0\), \(\omega_2=0.5\) \(\omega_1=0.5\) compared to other combinations with \(\omega_3=1.0\). Other combinations of the coefficients also result in a similar score in LAION-AesPredv2 (e.g. \(\omega_1=1.0\), \(\omega_2=0.7\) \(\omega_1=0.2\)), however, relying solely on LAION-AesPredv2 does not fully capture the method's aesthetic quality, as it is being optimized by the method.


\begin{table*}[htbp!]
  \centering
  \resizebox{\textwidth}{!}{%
    \begin{tabular}{l c c c c c c c c c}
      \multicolumn{10}{c}{} \\ \hline
      \multirow{3}{*}{}
        & $\omega_1\!=\!1.0$ & $\omega_1\!=\!1.0$ & $\omega_1\!=\!1.0$ & $\omega_1\!=\!1.0$ & $\omega_1\!=\!0.7$ & $\omega_1\!=\!1.0$ & $\omega_1\!=\!1.0$ & $\omega_1\!=\!1.0$ & $\omega_1\!=\!1.0$ \\
        & $\omega_2\!=\!1.0$ & $\omega_2\!=\!1.0$ & $\omega_2\!=\!1.0$ & $\omega_2\!=\!0.7$ & $\omega_2\!=\!1.0$ & $\omega_2\!=\!0.2$ & $\omega_2\!=\!0.5$ & $\omega_2\!=\!0.7$ & $\omega_2\!=\!1.0$ \\
        & $\omega_3\!=\!0.7$ & $\omega_3\!=\!0.5$ & $\omega_3\!=\!0.2$ & $\omega_3\!=\!1.0$ & $\omega_3\!=\!1.0$ & $\omega_3\!=\!0.2$ & $\omega_3\!=\!0.5$ & $\omega_3\!=\!0.2$ & $\omega_3\!=\!1.0$ \\ \hline
      LAION-AesPredv2 $\uparrow$
        & 0.60 $\pm$ 0.0021 & 0.60 $\pm$ 0.0024 & 0.59 $\pm$ 0.0022 & 0.60 $\pm$ 0.0020 & 0.59 $\pm$ 0.0023 & 0.60 $\pm$ 0.0020 & \textbf{0.60} $\pm$ 0.0019 & 0.60 $\pm$ 0.0018 & 0.60 $\pm$ 0.0022 \\
      HPSv2 $\uparrow$
        & 0.26 $\pm$ 0.0002 & 0.26 $\pm$ 0.0002 & 0.26 $\pm$ 0.0003 & 0.26 $\pm$ 0.0003 & 0.26 $\pm$ 0.0003 & 0.26 $\pm$ 0.0002 & \textbf{0.28} $\pm$ 0.0002 & 0.26 $\pm$ 0.0002 & 0.27 $\pm$ 0.0003 \\
      CLIPScore $\uparrow$
        & 0.27 $\pm$ 0.0017 & 0.27 $\pm$ 0.0018 & 0.27 $\pm$ 0.0017 & 0.27 $\pm$ 0.0017 & 0.27 $\pm$ 0.0016 & 0.27 $\pm$ 0.0015 & \textbf{0.27} $\pm$ 0.0016 & 0.27 $\pm$ 0.0018 & 0.27 $\pm$ 0.0019 \\ \hline
    \end{tabular}%
  }
  \caption{Hyperparameter search for the objective function's terms on 150 random captions from the training set of Promptist. The combination $\omega_1=1.0$, $\omega_2=0.5$, $\omega_3=0.5$ achieves a higher HPSv2 in comparison to other combinations.}
  \label{tab:tab_1_sup}
\end{table*}

\begin{figure*}[htbp!]
  \centering   \includegraphics[width=1.0\linewidth]{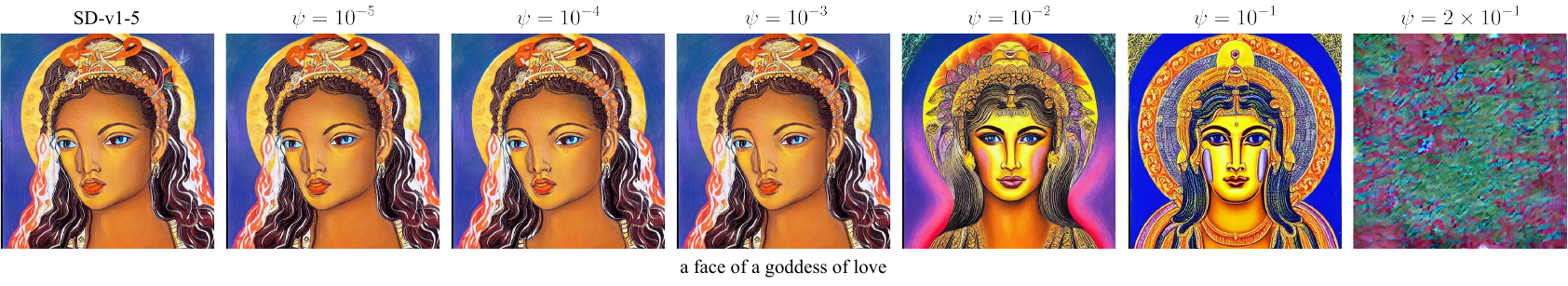}
   \caption{Hyperparameter search for the learning rate \(\psi\). \(\psi=10^{-2}\) enhances the aesthetic quality of the generated image and text-to-image alignment compared to SD-v-1-5 and results obtained with other learning rate values and does not diverge.}
   \label{fig:fig_2_sup}
\end{figure*}

\subsection{Hyper-parameter search for the Learning Rate}

\Cref{fig:fig_2_sup} provides results of the hyperparameter search on the learning rate \(\psi\) using the Adam optimizer with \(10\) optimization steps, compared to the baseline SD-v1-5. Low values of the learning rate (e.g.,  \(\psi=10^{-5}\) ) have a slight influence on the generated output, while high values of the learning rate (e.g., \(\psi=2\times10^{-1}\)) lead to divergence and corrupted output. A learning rate of \(\psi=10^{-2}\) demonstrates visual quality improvement over SD-v1-5 with enhanced text-to-image alignment. Therefore, \(\psi=10^{-2}\) is selected as the default value for the learning rate in the proposed PEO approach.

\section{Additional Qualitative Comparison with Baselines}
\label{sec:sup_3}

\begin{figure*}[htbp!]
  \centering   \includegraphics[width=1.0\linewidth]{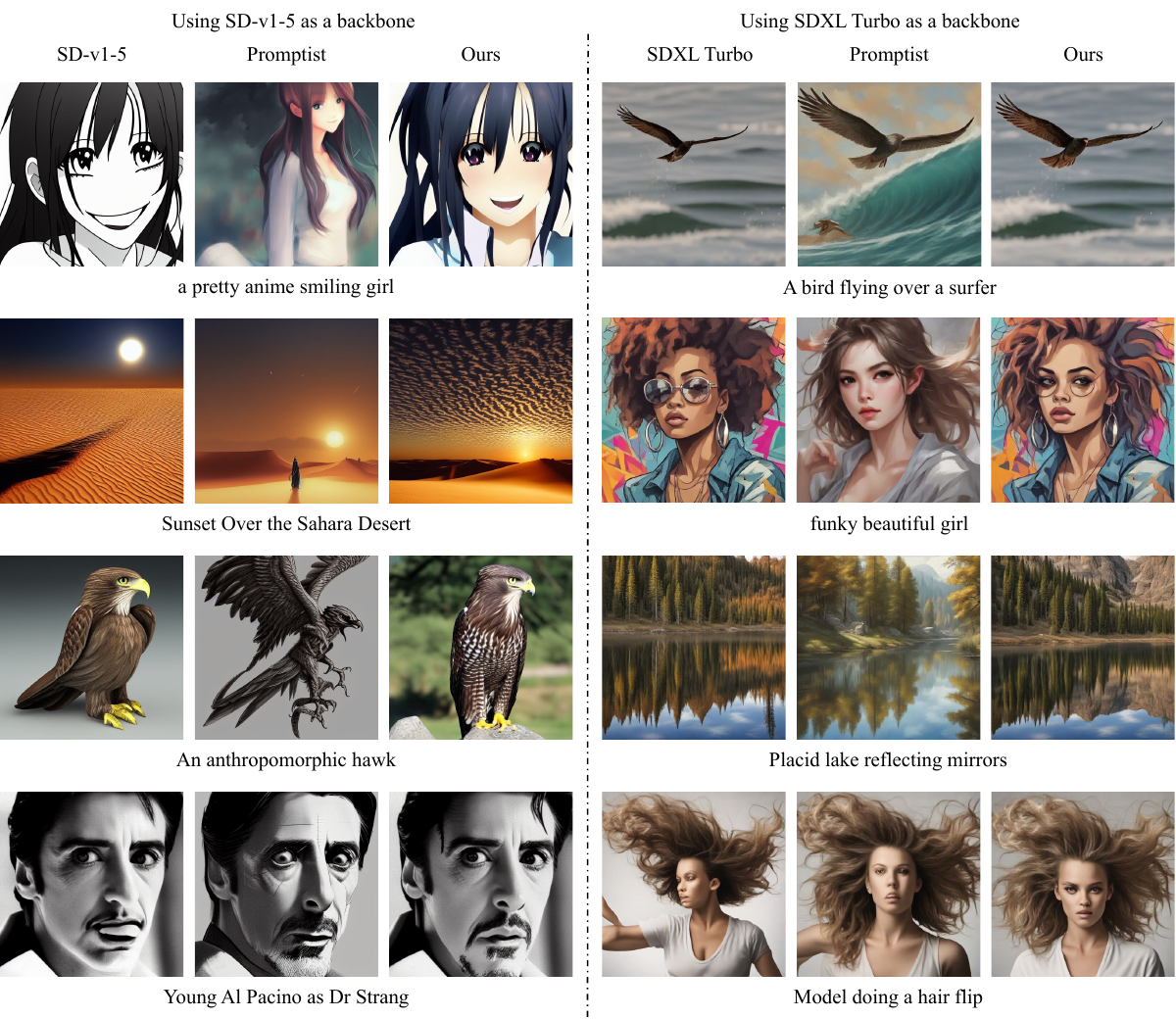}
   \caption{Qualitative comparison of PEO and baselines with SD-v1-5 and SDXL Turbo as backbones (Part 1). Our approach surpasses or matches the baseline in visual aesthetic quality, exhibits improved details, and better alignment with the style and main subject indicated in the original prompt.}
   \label{fig:fig_1_sup-part1}
\end{figure*}

\begin{figure*}[htbp!]
  \centering   \includegraphics[width=1.0\linewidth]{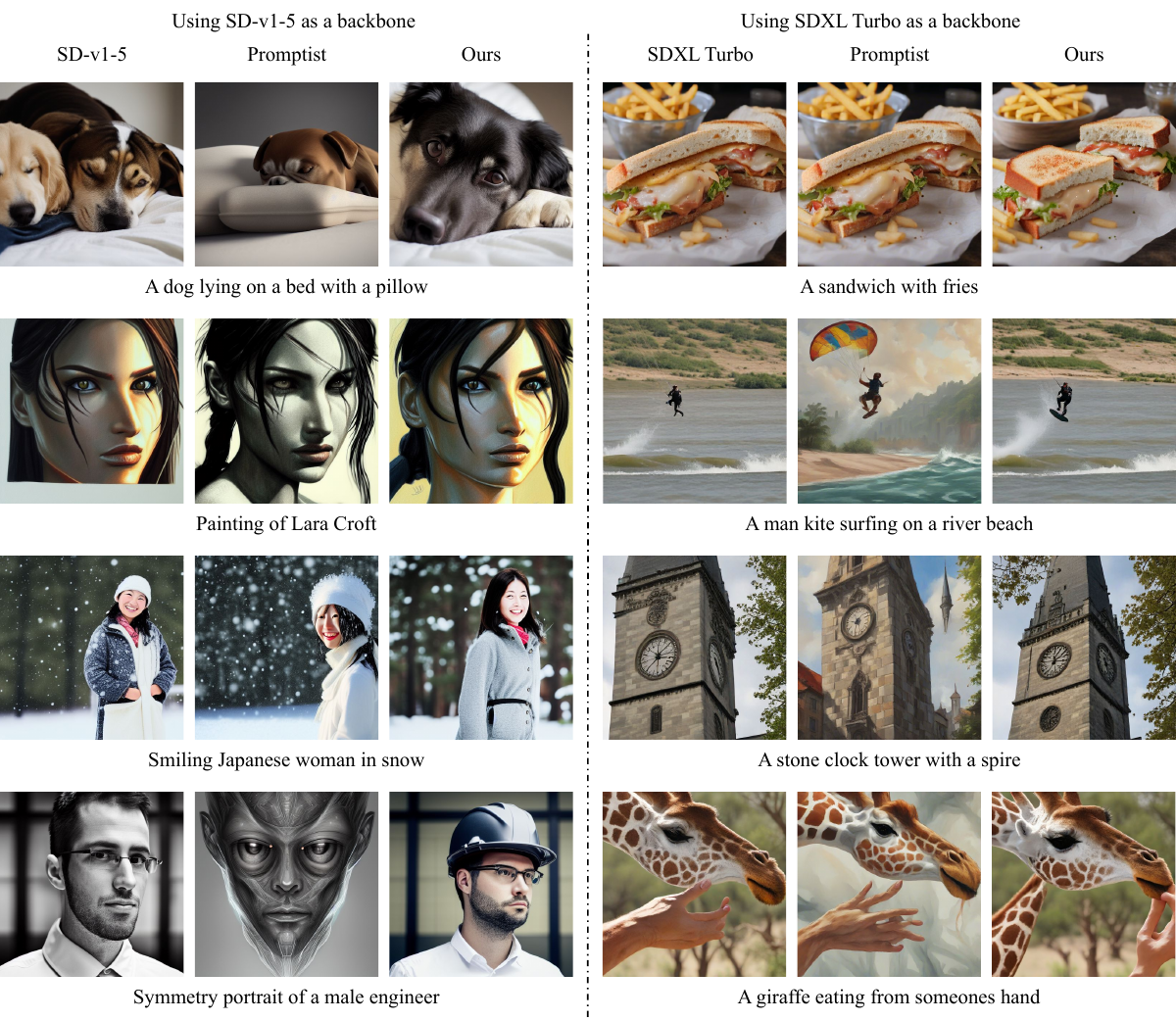}
   \caption{Qualitative comparison of PEO and baselines with SD-v1-5 and SDXL Turbo as backbones (Part 2). Our approach surpasses or matches the baseline in visual aesthetic quality, exhibits improved details, and better alignment with the style and main subject indicated in the original prompt.}
   \label{fig:fig_1_sup-part2}
\end{figure*}

\begin{figure*}[htbp!]
  \centering   \includegraphics[width=1.0\linewidth]{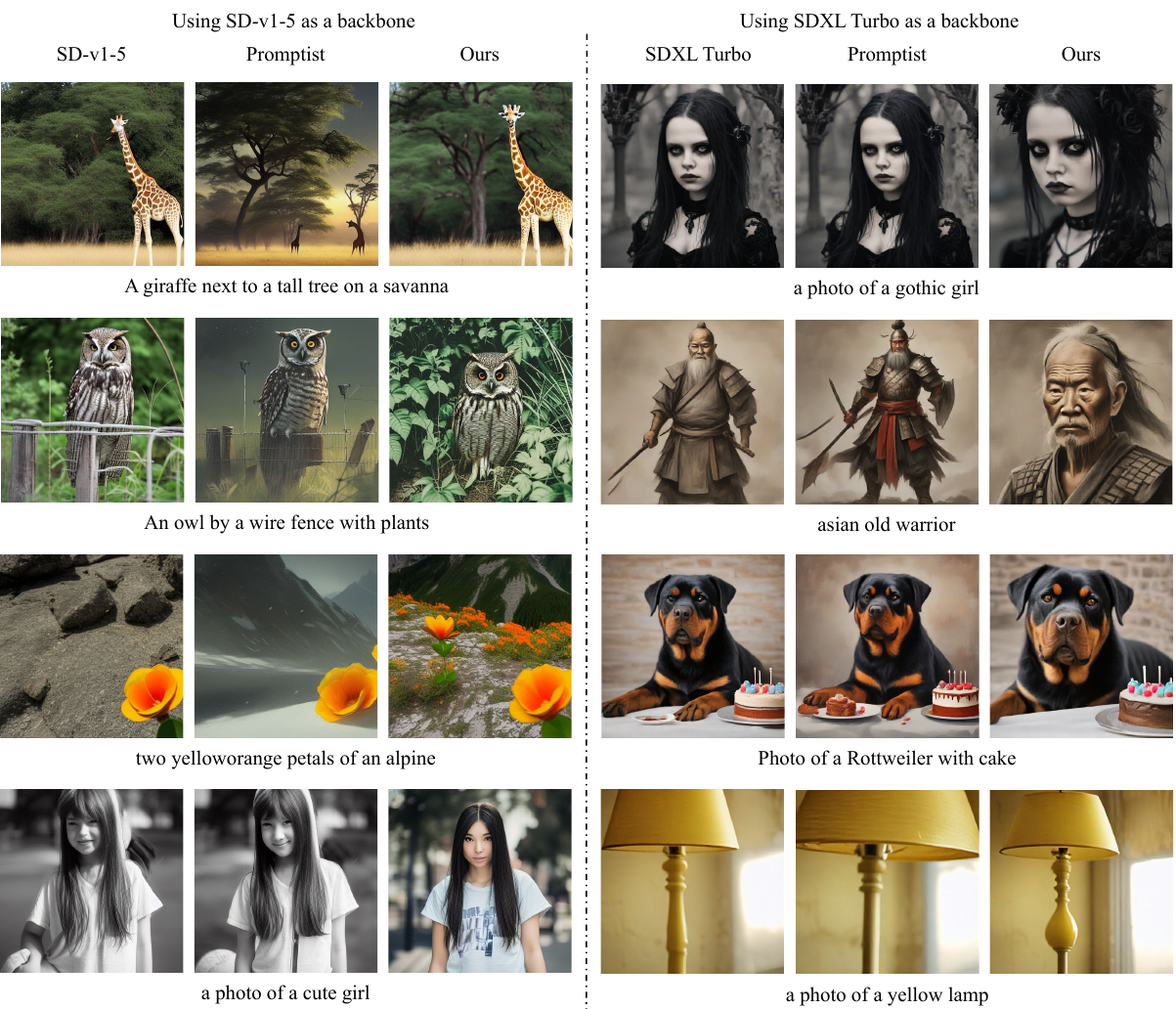}
   \caption{Visual comparison of PEO and baselines with SD-v1-5 and SDXL Turbo as backbones (Part 3). Our approach surpasses or matches the baseline in visual aesthetic quality, exhibits improved details, and better alignment with the style and main subject indicated in the original prompt.}
   \label{fig:fig_1_sup-part3}
\end{figure*}

Additional qualitative results comparing the proposed PEO approach with the baseline text-to-image diffusion models SD-v1-5 and SDXL Turbo and the prompt adaptation technique,
Promptist using SD-v1-5 and SDXL Turbo as backbones are shown in \cref{fig:fig_1_sup-part1,fig:fig_1_sup-part2,fig:fig_1_sup-part3}. The original text prompt is shown below each result. PEO demonstrates improved aesthetic quality and superior visual fidelity compared to baseline methods, achieving better alignment with the style and subject of the original prompt (e.g., “\underline{funky} beautiful girl” in \cref{fig:fig_1_sup-part1} and “a photo of a \underline{gothic} girl” in \cref{fig:fig_1_sup-part3}). Moreover, PEO produces more realistic results than Promptist, which often generates dreamlike outputs (e.g., “An anthropomorphic hawk” in \cref{fig:fig_1_sup-part1} and “Symmetry portrait of a male engineer” in \cref{fig:fig_1_sup-part2}). Our method also shows better alignment with the original prompt compared to the baselines (e.g., “Young Al Pacino as Dr. Strange” in \cref{fig:fig_1_sup-part1}), likely due to the PPT term in the objective function of PEO. Thus, our method enhances the aesthetic quality of the generated images without affecting text-to-image alignment.

\end{document}